\documentclass[10pt,twocolumn,letterpaper]{article}

\usepackage{cvpr}              % To produce the CAMERA-READY version
\usepackage[accsupp]{axessibility} % Improves PDF readability for those with disabilities.
\usepackage[dvipsnames]{xcolor}

\definecolor{cvprblue}{rgb}{0.21,0.49,0.74}
\usepackage[pagebackref,breaklinks,colorlinks,citecolor=cvprblue]{hyperref}

\usepackage{algorithm}
\usepackage{algorithmic}

\usepackage{amsfonts}
\usepackage{amsmath}
\usepackage{subcaption}
\usepackage{multirow}
\usepackage{booktabs}
\usepackage{xcolor}

\newcommand\blfootnote[1]{%
  \begingroup
  \renewcommand\thefootnote{}\footnote{#1}%
  \addtocounter{footnote}{-1}%
  \endgroup
}

\def\paperID{7476} % *** Enter the Paper ID here
\def\confName{CVPR}
\def\confYear{2024}

\title{Diversity-aware Channel Pruning for StyleGAN Compression}

\author{Jiwoo Chung, Sangeek Hyun, Sang-Heon Shim, Jae-Pil Heo$^{\ast}$ \\
Sungkyunkwan University\\
{\tt\small {\{wldn0202, hsi1032, ekzmwww, jaepilheo\}@g.skku.edu}}}

\usepackage{lipsum}

\begin{document}
\maketitle
\begin{abstract}
StyleGAN has shown remarkable performance in unconditional image generation. However, its high computational cost poses a significant challenge for practical applications. Although recent efforts have been made to compress StyleGAN while preserving its performance, existing compressed models still lag behind the original model, particularly in terms of sample diversity. To overcome this, we propose a novel channel pruning method that leverages varying sensitivities of channels to latent vectors, which is a key factor in sample diversity. Specifically, by assessing channel importance based on their sensitivities to latent vector perturbations, our method enhances the diversity of samples in the compressed model. Since our method solely focuses on the channel pruning stage, it has complementary benefits with prior training schemes without additional training cost. Extensive experiments demonstrate that our method significantly enhances sample diversity across various datasets. Moreover, in terms of FID scores, our method not only surpasses state-of-the-art by a large margin but also achieves comparable scores with only half training iterations. Codes are available at \href{https://github.com/jiwoogit/DCP-GAN}{github.com/jiwoogit/DCP-GAN}.

\blfootnote{
$^\ast$ Corresponding author
}
\end{abstract}

\vspace{-0.1cm}
\section{Introduction}
\label{introduction}
\begin{figure*}[t!]
    \centering
     \includegraphics[width=0.9\textwidth]{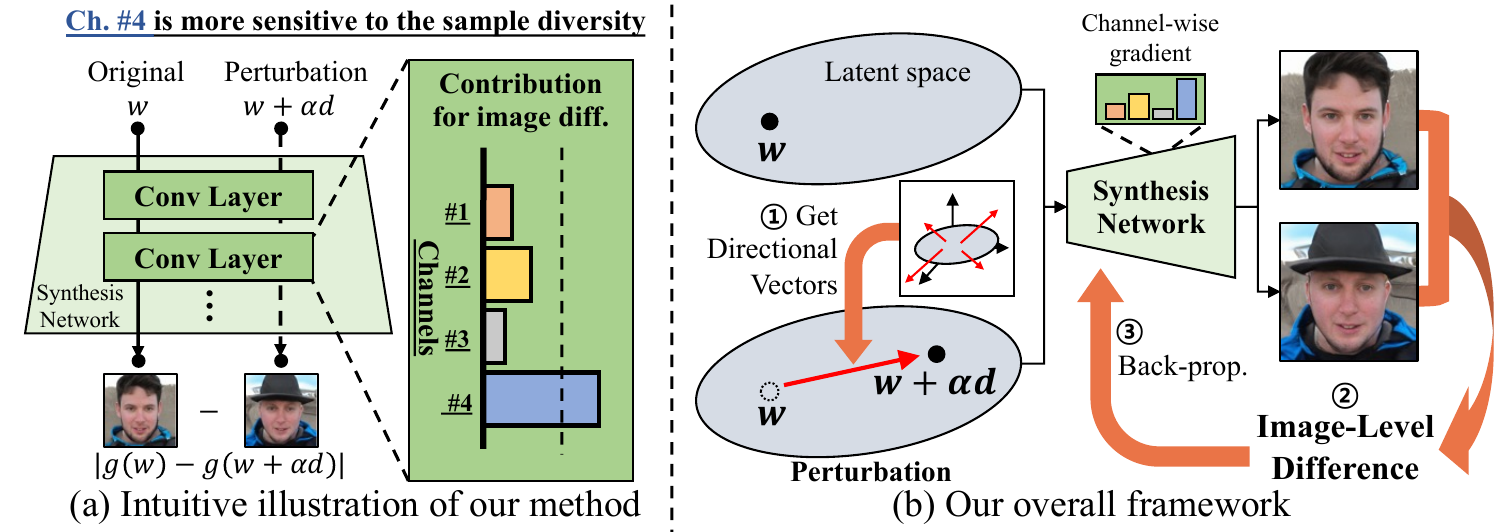}
     \vspace{-0.1cm}
     \caption{
         \textbf{(a) Intuitive illustration of our method.}
         We compare four channels~(Ch.~\#1, 2, 3, 4) by evaluating their responses when we pass the same latent vector $w$ and its perturbed counterpart $(w+\alpha d)$. By investigating the contribution of each channel to resulting image difference, we determine the sensitivity of channels to the latent perturbation. In this example, Ch.~\#4 is highly sensitive to the perturbation, while Ch.~\#1, 2, 3 exhibit low sensitivity. Consequently, in terms of preserving sample diversity, Ch.~\#4 is suitable for retaining.
         \textbf{(b) Our overall framework.}
         We aim to assess the contribution of each channel to the sample diversity by measuring its sensitivity to latent vector perturbation. In detail, 1) we sample a directional vector for the perturbation, 2) we compute the image-level difference caused by the latent vector perturbation, and 3) we calculate channel-wise gradient magnitudes induced by the difference image. The channel-wise sensitivity to the sample diversity is determined by its gradient magnitudes.
         As a result, we can estimate the channel-wise sensitivity against diversity. 
     }
     \label{motivation}
     \vspace{-0.2cm}
\end{figure*}

Thanks to recent progress in generative artificial intelligence, there have been numerous techniques in the research field of image generation~\cite{rombach2022high, esser2021taming, brock2018large, shim2023towards, hyun2023frequency, lee2022ggdr, park2019semantic, wang2022diffusion, sauer2022stylegan, kim2021exploiting}.
Especially for Generative Adversarial Networks~(GANs)~\cite{goodfellow2014generative}, StyleGAN~\cite{karras2019style,karras2020analyzing,karras2021alias} is one of the most successful approaches and leads to the development of diverse applications of GANs including image editing~\cite{harkonen2020ganspace,patashnik2021styleclip,zhu2020domain}, super resolution~\cite{richardson2021encoding}, and even 3D generation~\cite{gu2021stylenerf,shi2021lifting,chan2021pi}. 
However, due to their high computation costs, the deployment of these applications on the edge devices such as mobile phones and embedded systems is still challenging and remains a significant research problem.
To handle this challenge, the research area of StyleGAN compression~\cite{liu2021content,xu2022mind} is recently introduced.

Most of prior StyleGAN compression techniques including CAGAN~\cite{liu2021content} and StyleKD~\cite{xu2022mind} have two stages: (1) channel-pruning and (2)  distillation stages.
In detail, the channel-pruning stage initializes the compressed model~(student) from the pre-trained network~(teacher) by selectively removing channels.
Subsequently, in the distillation stage, channel-pruned student model undergoes further training by adversarial objectives and knowledge distillation from the teacher model if needed.

The primary goal of the GAN compression is to maintain the diversity and fidelity of generated images from the teacher model. However, we observe that the compressed generator often struggles to preserve the diversity compared to teacher generator~(i.e. low recall as reported in Table.~\ref{main-table}). Specifically, in the case of StyleKD, although it retains the latent space and mapping network of the teacher model during the pruning stage, the synthesis network of student model hardly preserves the diversity (recall) in the latent-to-image rendering process.
We hypothesize that such degradation in diversity is caused by improper initialization of the synthesis network~(e.g. random initialization or inadequate pruning method).
Hence, in this paper we focus on development of an appropriate channel pruning scheme for synthesis network toward preserving the diversity of teacher.

The diversity of the synthesis network stems from the variations observed in images generated from different latent signals. Our hypothesis is that the channel-wise behavior differs for different latent vectors, implying that each channel contributes to the diversity of generated images differently. For instance, channels associated with common characteristics in the dataset (e.g., the number of eyes in a facial dataset) may not be sensitive to the latent vector and its impact on diversity, despite their significant contribution to the generated images. On the other hand, channels related to semantic attributes exhibit distinct behavior with respect to the latent vector. Furthermore, the common characteristics within a dataset can be highly recoverable during further training even if their relevant channels are pruned. Therefore, to achieve a compressed student model that maintains a highly similar generative distribution to the teacher model with fewer channels, it is reasonable to prioritize the preservation of channels that are sensitive to the latent vector. 

In this paper, we present a channel-pruning method that considers the sensitivity of each channel to latent vector perturbations. Specifically, we investigate gradients induced by the image-level difference between two generated samples: one from the original latent vector and the other from its perturbated counterpart, as shown in Fig.~\ref{motivation}. The intuition behind this is that a larger gradient magnitude for a generator’s parameter indicates a higher sensitivity to the latent vector perturbation. Building on this primitive idea, we define a channel-wise importance score by aggregating the gradient magnitudes of parameters within the channel for diverse latent vectors and their perturbations. This score serves as a measure of each channel’s contribution to the sample diversity. Based on these importance scores, we prune channels accordingly.

We conduct extensive evaluations in various datasets including FFHQ, LSUN-Church, and LSUN-Horse. The experimental results demonstrate that our compressed model not only achieves state-of-the-art performance but also preserves significantly higher sample diversity. 
Moreover, our method reaches the previous state-of-the-art FID score with only half the number of training iterations.
\vspace{-0.1cm}
\section{Related Work}
\label{related_work}
One of the network compression techniques, channel pruning~\cite{molchanov2016pruning, sun2017meprop, zhuang2018discrimination, liu2019channel, lee2020channel, liu2021discrimination, zhang2022wavelet} proves highly effective in significantly reducing both memory usage and computational expenses. For example, MeanGrad~\cite{liu2019channel} aims to preserve model performance despite pruning in network parameters and evaluates the importance of convolution weights using the mean gradient criterion in classification tasks.

In generation tasks, various methods~\cite{kang2022information, hu2023discriminator, ren2021online, li2021revisiting, li2022learning} have been also proposed to compress the generator network in both conditional and unconditional GANs to leverage its generative capabilities on devices with limited computational resources. One of the recent works, CAGAN~\cite{liu2021content} proposed a two-stage compression method for unconditional GANs. They first pruned the network parameters and then fine-tuned the pruned student model using knowledge distillation from the teacher model. Additionally, they proposed a content-aware pruning technique to preserve channels that are highly activated in salient regions of images.
More recently, StyleKD~\cite{xu2022mind} addressed the issue of output discrepancy between teacher and student models by maintaining the mapping network while randomly initializing the synthesis network with a reduced number of parameters.

Our approach revisits the pruning stage in compressing the generator of StyleGAN.
Specifically, we introduce a novel algorithm for selecting channels to be pruned based on their contributions to the sample diversity measured by the sensitivity to latent vector perturbations.
Additionally, our method requires no extra training costs during the pruning stage, as it operates without the need for additional supervision, such as a pre-trained semantic segmentation model, while CAGAN requires it.
In StyleKD, despite discarding all parameters in the synthesis network and utilizing latent directions to mimic the teacher model in the distillation stage, we observe that they still face challenges in maintaining sample diversity. In contrast, our proposed method directly retains diversity-aware parameters in the synthesis network while discarding diversity-unrelated parameters in the pruning stage. Thus, our approach effectively inherits the generative capabilities of the teacher network and is twice as fast as the baselines thanks to our enhanced initialization.

\vspace{-0.1cm}
\section{Method}
\label{method}
In this paper, we present a novel channel pruning strategy for compressing StyleGAN. we focus on preserving sample diversity by considering the relationship between the latent and image spaces of the GAN, which has been overlooked by previous methods in the pruning process. Before introducing details of our proposed method, we provide an overview of the typical pipeline of GAN compression. Subsequently, we describe our channel selection process.

\subsection{Preliminaries}
\label{preliminaries}

\begin{figure}[t!]
    \includegraphics[width=0.9\columnwidth]{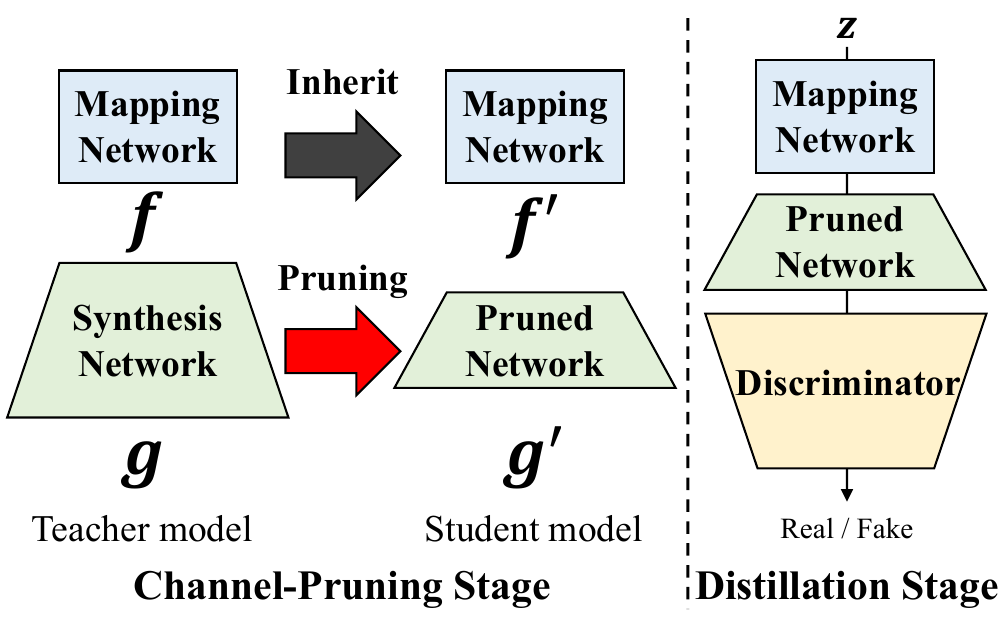}\centering
    \vspace{-0.2cm}
    \caption{
    \textbf{Overall compression framework.}
    Overall compression framework consists of two stages: channel pruning and distillation. Channel-pruning stage initializes a compact student model by pruning channels of a larger teacher model. Specifically for the StyleGAN2 architecture, pruning process usually focuses on reducing channels in the synthesis network, while retraining the mapping network. Distillation stage further trains student model with several training objectives such as adversarial and distillation losses. In this paper, we focus on the channel-pruning stage.
    }
    \label{overall_procedure}
    \vspace{-0.2cm}
\end{figure}

\noindent\textbf{Unconditional GAN Compression.}
As in Fig~\ref{overall_procedure}, GAN compression methods typically involve two networks: a teacher model (large and pretrained) and a student model (small). The goal is to train the student model with fewer parameters by distilling the knowledge from the pretrained teacher network, while maintaining performance as much as possible. This approach generally consists of two stages: (1) channel pruning and (2) distillation. In the channel pruning stage, the less important channels of the teacher model are pruned to initialize the student model. In the distillation stage, the student model is further trained using both adversarial learning and distillation losses from the teacher.

\vspace{0.2cm}
\noindent\textbf{Channel-Pruning Stage.}
The channel-pruning stage involves the estimation of channel importance within the generator network, followed by pruning of less important channels based on the determined importance scores. In the context of StyleGAN compression, since most of methods are built on the StyleGAN2 framework~\cite{karras2020analyzing}, we provide a detailed description of the pruning procedure. The generator architecture of StyleGAN2 consists of a mapping network, denoted as $f(\cdot):\mathcal{Z}\mapsto\mathcal{W}$, and a synthesis network, denoted as $g(\cdot):\mathcal{W}\mapsto\mathcal{I}$.
Here, $\mathcal{Z}\in\mathbb{R}^{512}$ represents the input noise space following a standard normal distribution $\mathcal{N}(0,1)$.
$\mathcal{W}\in\mathbb{R}^{512}$ corresponds to the intermediate latent space, and $\mathcal{I}\in\mathbb{R}^{h\times w \times 3}$ represents the image space.
In the case of StyleGAN2, the focus of pruning is primarily on reducing the channels within the synthesis network while maintaining the mapping network. This strategy is guided by two reasons. Firstly, the mapping network accounts for only a small portion of the parameters and floating-point operations (FLOPs) compared to the synthesis network, making the benefits of pruning the mapping network negligible. Secondly, previous work~\cite{xu2022mind} has observed that pruning the mapping network results in a significant drop in performance as it disrupts the consistency of output images between teacher and student models. Therefore, most of the existing research in this field has concentrated on developing techniques to prune the synthesis network. In alignment with this approach, our paper also focuses on pruning the synthesis network.

\subsection{Our Approach}
\label{our_method}
As discussed earlier, we observed a notable degradation in sample diversity in the compressed student model compared to the teacher network, even though the mapping network remains uncompressed.
We hypothesize that a critical factor in this decline is the lack of consideration for sample diversity in previous network initialization (or channel pruning).
Consequently, our primary goal is to address this issue by introducing a novel channel-pruning method aimed at enhancing the diversity of generated images.

To accomplish this, we introduce the concept of latent perturbation-induced gradients. These gradients reflect the sensitivity of each channel to perturbations in the latent vector. Since latent perturbations produce the image-level differences of generated samples, the gradients induced by these perturbations enable us to measure the contribution of each channel to the sample diversity. Based on these gradients, we define diversity-sensitive importance scores for each channel using various latent vectors and their perturbations. Finally, we select channels to be pruned according to their importance scores.

Note that, while we explain the method using a single convolution weight $W$ for simplicity, in practice, our pruning method is applied to all weights in convolutional layers.

\subsubsection{Latent Perturbation-induced Gradients}
\label{diverstiy-sensitive-grad}
In our context, sample diversity refers to the variations observed among images generated from different latent vectors. To assess the contribution of each channel to sample diversity, we examine the gradients induced by the image-level difference between two generated samples from the latent vector and its perturbation. Specifically, we consider the latent representations in the intermediate latent space $\mathcal{W}$ inherited from the teacher model.

To produce the perturbated latent vectors, we first establish a set of independent directional vectors $D$ in $\mathcal{W}$. The perturbation is accomplished by shifting the latent vector $w$ along a directional vector $d$ sampled from $D$ as $w+\alpha d$, where $\alpha$ is a scalar constant. Among a variety of feasible strategies to define $D$, we specifically employ principal components inspired by GANSpace~\cite{harkonen2020ganspace}. We also use a probabilistic sampling for the selection of directional vectors, with the selection probabilities proportional to the variance ratios corresponding principal components, since directions represent different degrees of semantic variation in this scheme. Note that, we experimentally found that our algorithm also works effectively even with a set of randomly drawn directional vectors~(i.e.~($d_{\text{rand}}\sim \mathcal{N}(0,1)$), and the performance gap between principal components and random vectors is not significant.

Given a latent vector $w$ and a sampled directional vector $d$, we compute the difference between two images $g(w)$ and $g(w+\alpha d)$ generated by the teacher generator $g(\cdot)$ based on $L_1$ distance as follows:
\begin{equation}
\label{eq_sem_loss}
\mathcal{L}_{\text{diff}}=\vert g(w)-g(w+\alpha d) \vert, \quad g(\cdot)\in \mathbb{R}^{h\times w \times 3},
\end{equation}
where $\vert\cdot\vert$ denotes the absolute value, and $h$ and $w$ are the height and width of the images, respectively.

To investigate the sensitivity of learnable parameters $W$ within a convolution layer of $g$ to the latent vector perturbation, we calculate gradients of weights by back-propagating $\mathcal{L}_{\text{diff}}$. These gradients are referred to as latent perturbation-induced gradients $\mathbf{G}_{\text{perturb}}$, and expressed as follows:
\begin{equation}
\label{eq_grad_norm}
\mathbf{G}_{\text{perturb}}=\vert{\frac{\partial \mathcal{L}_{\text{diff}}}{\partial W}}\vert, \quad \mathbf{G}_{\text{perturb}}\in\mathbb{R}^{c^{\text{in}}\times c^{\text{out}}\times 3\times 3},
\end{equation}
where $c^{\text{in}}$ and $c^{\text{out}}$ represent the number of input channels and output channels of $W$, respectively.

Intuitively, if a parameter has a larger gradient magnitude, the parameter is considered more sensitive to the difference between samples produced by latent perturbation, thereby contributing more to the sample diversity.

\subsubsection{Diversity-Sensitive Importance Score}
\label{direction-variacne-section}
Based on the latent perturbation-induced gradients $\mathbf{G}_{\text{perturb}}$, we introduce channel-wise diversity-sensitive importance score for pruning. 

One straightforward approach is to average the gradient magnitudes over the parameters associated with each channel for multiple latent vectors $w\sim f(z)$ and directions $d$.
We refer this scheme as the average-based importance score, expressed as follows:
\begin{equation}
\label{mean_eq}
S^{\mu}(c)= \Vert\mathbb{E}_z\mathbb{E}_d[\mathbf{G}_{\text{perturb}}]_c\Vert_1,S^{\mu}(c)\in\mathbb{R}^1,
\end{equation}
where $c\in[0,c^{\text{in}})$ denotes the channel index, $[\cdot]_c$ represents $c^{\text{th}}$ channel of $[\cdot]$, and $\mathbf{G}_{\text{perturb}}$ denotes the gradient magnitude. In practice, we compute $\mathbb{E}_d [ \cdot ]$ using $N$ directional vectors sampled from $D$.

If a specific latent vector significantly influences $[\mathbf{G}_{\text{perturb}}]_c$ regardless of its perturbations, the channel would receive a high importance score even if it is not actually sensitive to the perturbations. Consequently, the influence of such channels can greatly disturb desired channel selection.

To address this problem, we revise the average-based score by considering the individual effect of each latent vector.
Specifically, for each latent vector, we compute the average gradient magnitudes over $N$ perturbations and utilize them to penalize the induced gradients for the learnable parameters. These average gradient magnitudes for a latent vector are referred to as the gradient offset. As a result, our diversity-sensitive importance score for $c^{\text{th}}$ channel, $S^{\sigma}(c)$, is defined as follows:
\begin{equation}
\label{var_eq}
S^{\sigma}(c)= \Vert\mathbb{E}_z\mathbb{E}_d[[\mathbf{G}_{\text{perturb}}-\mathbb{E}_d[\mathbf{G}_{\text{perturb}}]]^2]_c\Vert_1, S^{\sigma}(c)\in\mathbb{R}^1,
\end{equation}
where $\mathbb{E}_d[\mathbf{G}_{\text{perturb}}]$ represents the gradient offset of the original latent vector $w$, and $[\cdot]^2$ denotes element-wise squaring. Once the diversity-sensitive importance scores are computed, we prune channels with low scores according to the pruning ratio $p_r$.

Aside from the calibration by the gradient offset, $S^{\sigma}(c)$ is equivalent to the variance of $\mathbf{G}_{\text{perturb}}$ over perturbations. This approach further reduces the impact of small noisy values and grants greater importance to channels that are highly sensitive to a particular direction in the latent space. Such directions can be highly related to specific semantic attributes. Intuitively, it is crucial to retain channels that are highly sensitive to directional vectors in the latent space relevant to certain semantic attributes, such as glasses or age.

\subsection{Objective Functions for Distillation Stage}
\label{objectives}
After establishing the initial student model through a channel-pruned teacher generator, we fine-tune the student model using adversarial and knowledge distillation objectives. As our primary focus in this paper lies on the pruning algorithm, we simply follow the training scheme of StyleKD~\cite{xu2022mind} incorporating four objectives: $\mathcal{L}_{\text{GAN}},\mathcal{L}_{\text{rgb}},\mathcal{L}_{\text{lpips}}$, and $\mathcal{L}_{\text{LD}}$.

$\mathcal{L}_{\text{GAN}}$ denotes the minimax objective for adversarial training. Meanwhile, $\mathcal{L}_{\text{rgb}}$ and $\mathcal{L}_{\text{lpips}}$ strive to replicate teacher-generated images in image and feature spaces, respectively. $\mathcal{L}_{\text{LD}}$, a latent-direction-based relation distillation loss, aims to align the student model's feature similarity matrix to the teachers. More details on these objectives can be found in StyleKD~\cite{xu2022mind}.
As a result, our final training objective function, $\mathcal{L}_{\text{final}}$, is given as follows:
\begin{equation}
    \mathcal{L}_{\text{final}}= \lambda_{\text{GAN}}\mathcal{L}_{\text{GAN}}+\lambda_{\text{rgb}}\mathcal{L}_{\text{rgb}}+\lambda_{\text{lpips}}\mathcal{L}_{\text{lpips}}+\lambda_{\text{LD}}\mathcal{L}_{\text{LD}}, 
\end{equation}
where $\lambda_{\text{GAN}},\lambda_{\text{rgb}},\lambda_{\text{lpips}}$, and $\lambda_{\text{LD}}$ balance loss terms.
\section{Experiment}
\label{experiment}

\subsection{Experimental Setup}
\noindent\textbf{Baselines.}
Our method is compared with recent StyleGAN compression methods, CAGAN~\cite{liu2021content} and StyleKD~\cite{xu2022mind}. For a fair comparison, we further train CAGAN with the advanced objective function ($\mathcal{L}_{\text{LD}}$) introduced in StyleKD, which we denote as CAGAN$^*$. Since our method employs StyleKD's training scheme, comparing CAGAN$^*$, StyleKD, and our method is equitable. Since the pretrained StyleKD models are not available, we retrain StyleKD using the official repository for unreported metrics and datasets, referring to this retrained model as StyleKD$^*$. For training details of CAGAN, we apply a facial mask as a content mask when training CAGAN$^*$ on the FFHQ dataset. However, on other datasets where content definition is challenging, we use a uniform mask, assigning a value of one to all pixels.

\vspace{0.2cm}
\noindent\textbf{Evaluation Metrics.}
We validate our pruning method using various quantitative metrics. In addition to the most popular Fr\'echet Inception Distance~(FID)~\cite{heusel2017gans}, we also measure Precision and Recall~(P\&R)~\cite{kynkaanniemi2019improved} to assess the quality and diversity of generated samples separately. For FID calculation, we use 50K real and 50K fake samples for each dataset. For P\&R, we use all real samples in the dataset and 50K fake samples for FFHQ and LSUN Church. For the LSUN Horse dataset, we use 200K real samples due to the high computational cost associated with complete 1M real samples.

To evaluate the projection capability of the compressed model, we employ the projection method of StyleGAN2, which projects real images into the $\mathcal{W}$ space using noise regularization and reconstruction losses such as MSE and LPIPS, over 1,000 iterations. We use the Helen-Set55 dataset for this evaluation, following the practice of CAGAN. We report the averaged MSE and LPIPS for all pairs of real and projected images in the dataset.

\vspace{0.2cm}
\noindent\textbf{Implementation Details.}
We follow the implementation details provided in StyleKD~\cite{xu2022mind}. 
For instance, all experiments are conducted with a pruning ratio of 0.7~(i.e. $p_r=0.7$) and trained for 450K iterations.
Also, in the default setting, we set the hyperparameters $N$ and $\alpha$ as 10 and 5, except for LSUN Horse dataset where we set $\alpha$ as 10.
We employ non-saturating loss~\cite{NIPS2014_5ca3e9b1} with $R_1$ gradient penalty~\cite{karras2020analyzing} and update a model through Adam optimizer~\cite{kingma2014adam} with $\beta_1=0.0$ and $\beta_2=0.99$. The learning rate is set to 2e-3 for both the generator and discriminator. We use a batch size of 16 and consistently employ 4 GPUs for training. Lastly, the loss balancing constants are set to $\lambda_{\text{GAN}}=1$, $\lambda_{\text{rgb}}=3$, $\lambda_{\text{lpips}}=3$, and $\lambda_{\text{LD}}=30$.

\subsection{Quantitative Results}

\begin{table*}[t]
\small
  \caption{
    \textbf{Quantitative results on StyleGAN2.}
    Comparison with GAN compression baselines in FFHQ, LSUN Church, and LSUN Horse datasets. ‘Param.’ refers to the number of parameters in a generator and `FLOPs' denotes the number of floating-point operations. `P' and `R' represent precision and recall metrics, respectively. `MSE' and `LPIPS' indicate the distance between pairs of real and inverted samples based on the StyleGAN2 projection method. `CAGAN$^*$' denotes CAGAN trained with the objective of StyleKD. `StyleKD$^*$' refers to a retrained model using the official code due to the unavailability of published pretrained models. `$\psi$' denotes the parameter for the truncation trick, which adjusts the trade-off between sample diversity and fidelity. We adjust `$\psi$' to evaluate precision at the highest recall achieved by baselines.
    We have omitted `Param.' and `FLOPs' in the LSUN Church and LSUN Horse datasets, as they are the same as FFHQ-256. Reported FIDs without ours, CAGAN$^*$, and StyleKD$^*$ are taken from StyleKD.
    }
    \vspace{-0.1cm}
  \label{main-table}
  \centering
\setlength\tabcolsep{5.3pt}
\begin{tabular}{c|l||cc|ccc|cc}\toprule
Dataset & Method & Param. &FLOPs &FID$\downarrow$ &P$\uparrow$ &R$\uparrow$ &MSE$\downarrow$ &LPIPS$\downarrow$ \\\midrule \midrule 
\multirow{9}{*}{FFHQ-256} & Teacher & 30.0M &45.1B &4.5 &0.603 &0.482 &0.0292 &0.2065 \\ \cmidrule{2-9}
& Scratch & 5.6M &4.1B &9.79 &N/A &N/A &N/A &N/A \\
& GS~\cite{wang2020gan} & N/A &5.0B &12.4 &N/A &N/A &N/A &N/A \\
& CAGAN~\cite{liu2021content} & 5.6M &4.1B &7.9 &0.703 &0.295 &0.0316 &0.2141 \\
& CAGAN$^*$~\cite{liu2021content} & 5.6M &4.1B &7.41 &0.713 &0.309 &0.0301 &0.2113 \\
& StyleKD~\cite{xu2022mind} & 5.6M &4.1B &7.25 &N/A &N/A &N/A &N/A \\
% StyleKD* &5.6M &4.1B &7.47 &0.710 &0.306 &0.0121 &0.1298 \\
& StyleKD$^*$~\cite{xu2022mind} & 5.6M &4.1B &7.47 &0.710 &0.306 &0.0307 &0.2101 \\ \cmidrule{2-9}
& \textbf{Ours} & 5.6M &4.1B &\textbf{6.35} &0.706 &\textbf{0.339} &\textbf{0.0293} &\textbf{0.2082} \\
& \textbf{Ours}~($\psi=0.95$) & 5.6M &4.1B &6.86 &\textbf{0.732} &0.312 & N/A &N/A \\\midrule \midrule
\multirow{5}{*}[-0.5em]{FFHQ-1024} & Teacher &49.1M &74.3B &2.7 &0.688 &0.492 &0.0275 &0.1916 \\ \cmidrule{2-9} %\midrule
& GS~\cite{wang2020gan} & N/A &23.9B &10.1 &N/A &N/A &N/A &N/A \\ 
& CAGAN~\cite{liu2021content} & 9.2M & 7.0B &7.6 &\textbf{0.721} &0.277 &0.0358 &0.2099 \\
& StyleKD~\cite{xu2022mind} & 9.2M & 7.0B &7.19 &N/A &N/A &N/A &N/A \\ \cmidrule{2-9} %\midrule
& \textbf{Ours} & 9.2M & 7.0B &\textbf{5.80} &0.676 &\textbf{0.378} &\textbf{0.0328} &\textbf{0.2082} \\
\bottomrule
% \bottomrule
\end{tabular}
\setlength\tabcolsep{3.2pt}
\begin{tabular}{c|l||ccc|c|l||ccc} \toprule
Dataset & Method &FID$\downarrow$ &P$\uparrow$ & R$\uparrow$ & Dataset & Method &FID$\downarrow$ &P$\uparrow$ &R$\uparrow$ \\\midrule \midrule
\multirow{4}{*}[-1.0em]{\shortstack{LSUN \\ Church}} &Teacher & 3.97 &0.599 &0.388 & \multirow{4}{*}[-1.0em]{\shortstack{LSUN \\ Horse}} &Teacher & 4.50 &0.541 &0.442 \\ \cmidrule{2-5} \cmidrule{7-10} %\midrule
& CAGAN$^*$~\cite{liu2021content} & 5.09 &0.598 &0.352 & &CAGAN$^*$~\cite{liu2021content} &5.73 &0.589 &0.335 \\
& StyleKD$^*$~\cite{xu2022mind} & 6.10 &\textbf{0.611} &0.302  & &StyleKD$^*$~\cite{xu2022mind} &5.69 &0.589 &0.325 \\ \cmidrule{2-5} \cmidrule{7-10} %\midrule
& \textbf{Ours} &\textbf{4.87} &0.584 &\textbf{0.372}  & &\textbf{Ours} &\textbf{5.11} &0.581 &\textbf{0.356} \\
&\textbf{Ours}~($\psi=0.97$) &5.07 &0.610 &0.349  & &\textbf{Ours}~($\psi=0.95$) &5.22 &\textbf{0.610} &0.337 \\
\bottomrule
\end{tabular}
\vspace{-0.3cm}
\end{table*}

\begin{figure}[t!]
    \centering
     \includegraphics[width=1.0\columnwidth]{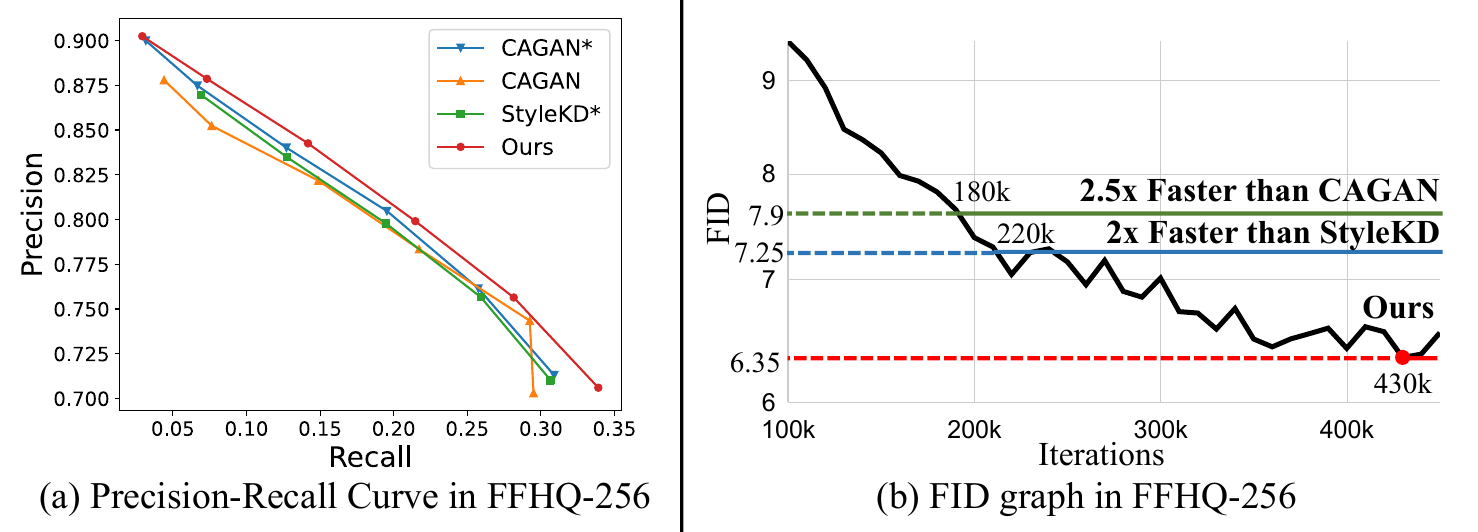}
     \vspace{-0.5cm}
     \caption{
         \textbf{(a) Precision-Recall Curve.} By adjusting the truncation trick parameter $\psi$ within the range [0.5, 1.0] with step size 0.1, we visualize the Precision-Recall curve of the proposed method and baselines. We validate that ours surpasses baseline methods with every range of precision and recall.
         \textbf{(b) FID w.r.t Training Iterations.} We visualize FID curve during training and validate that the proposed method achieves the previous state-of-the-art FIDs only with 2$\times$ fewer iterations.
     }
     \label{pr_graph}
     \vspace{-0.4cm}
\end{figure}

\noindent\textbf{Image Generation Performance.}
We validate the effectiveness of the proposed method through various generation performance metrics including FID and P\&R in diverse datasets such as FFHQ, LSUN Church, and LSUN Horse.
In terms of FID, the proposed method consistently outperforms state-of-the-art baselines on all tested datasets, as reported in Tab.~\ref{main-table}.
For example, our model achieves a FID score of 5.80 in FFHQ-1024, improving 1.8 and 1.39 points over CAGAN and StyleKD, respectively.
These consistent FID score improvements demonstrate the superiority of the distribution-matching capability of the proposed method.

For the recall metric directly related to the sample diversity, our method achieves the highest scores in all tested datasets.
For instance, our method significantly outperforms CAGAN in terms of recall score in FFHQ-1024 (0.277 vs 0.378), as well as obtains a value of 0.372 in LSUN Church, which is almost comparable with the teacher (0.016 points gap).

For the precision metric, the proposed method shows a slightly lower precision compared to the baseline, due to the precision-recall trade-off. Since the recall can be leveraged to improve precision through a truncation trick, we additionally conduct experiments by adjusting `$\psi$' to evaluate precision at the highest recall achieved by baselines.
In this setting, we observe that our method also surpasses baselines in terms of precision at the comparable recall.
For example, by adopting a truncation parameter of $\psi=0.95$ in FFHQ-256 and LSUN Horse datasets, our method provides higher precision, recall, and even FID scores compared to the baselines. 
In the case of LSUN Church, with a truncation parameter of $\psi=0.97$, we achieve comparable precision to StyleKD while providing 0.047 points higher recall score.
Similarly, compared to CAGAN, ours has comparable scores in the aspect of recall but achieves higher precision with a reasonable margin.
Note that, without the truncation trick, the proposed method substantially outperforms the baseline models in terms of diversity. For further validation, we additionally report the precision-recall curve in the FFHQ-256 dataset by adjusting the parameter $\psi$ in Fig.~\ref{pr_graph}-(a) within a range of [0.5, 1.0]. As shown, ours provides higher precision and recall compared to the baselines across the ranges.

To assess the applicability of our method with other network architectures, we conduct evaluations with StyleGAN3.
Thus, we reproduce scores of baselines at the same training settings as ours and focus on relative improvements in this experiment. 
Specifically, we employ official codes of StyleGAN3 and train the generator until the discriminator see 10M of real images.
As in Tab~\ref{stylegan3}, our model achieves a FID score of 8.40 and improves 5.85 and 1.17 points over the CAGAN and StyleKD, respectively. In terms of the recall metric, our method also surpasses the baselines. A comparative analysis of metric scores between baselines and our approach further highlights our model's superiority.

\vspace{0.2cm}
\noindent\textbf{Training Converged Speed Ability.}
Moreover, we visualize the FID curve with respect to the training iteration of our method in Fig.~\ref{pr_graph}-(b).
We observe that ours reach the best FID scores of previous state-of-the-art models at much fewer iterations.
For instance, ours achieve comparable FID scores only with 2.5$\times$ and 2$\times$ fewer iterations compared to the CAGAN and StyleKD, respectively.
We believe that our enhanced initialization of the synthesis network through our channel pruning strategy significantly improves the generation performance during the early stages of training, and allows the model for further improvement throughout the entire training process.

\vspace{0.2cm}
\noindent\textbf{Image Projection Ability.}
For evaluation of the image projection ability, we perform experiments on FFHQ dataset using MSE and LPIPS metrics.
For both FFHQ-256 and FFHQ-1024 datasets, we validate that the proposed method surpasses baselines in terms of both MSE and LPIPS metrics.
Interestingly, we observe that ours reports MSE of 0.0293 which is a comparable score with the teacher model, 0.0292, in FFHQ-256 dataset.
We believe that our pruning method brings a significant improvement in the generator's ability to capture the diverse characteristics of real images, and it allows precise projection of them.% real images.

\begin{figure*}[t!]
    \centering    
     \includegraphics[width=0.9\textwidth]{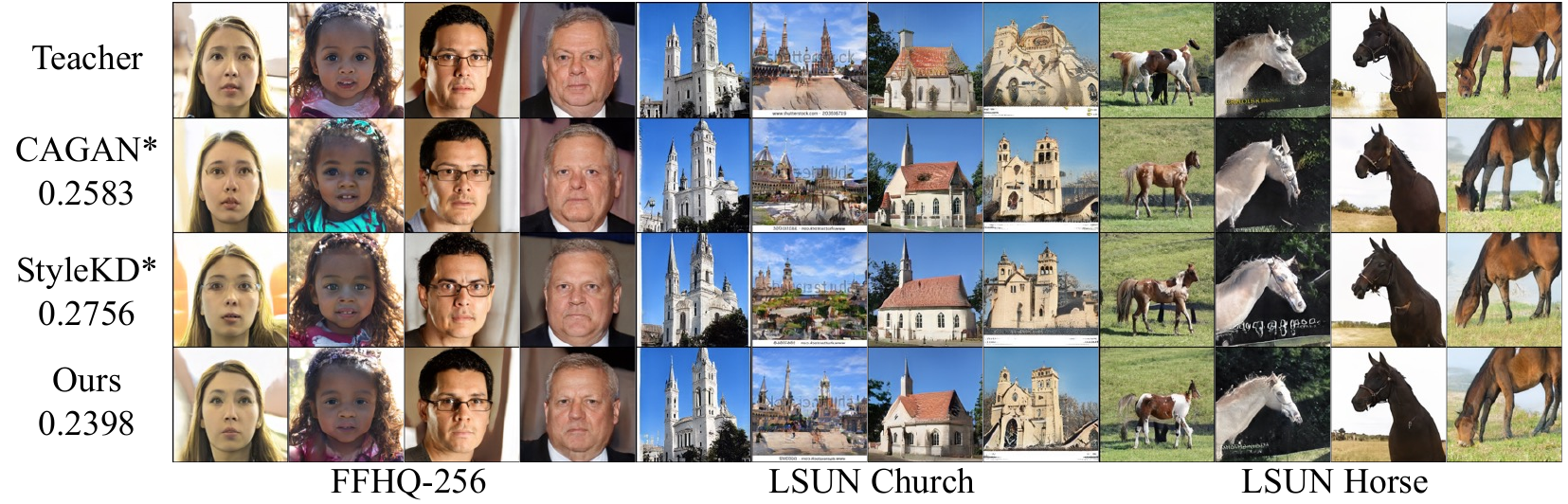}
     \vspace{-0.2cm}
     \caption{
         \textbf{Qualitative comparison with baselines on various datasets.}
         For qualitative comparison, we visualize our generated samples and baselines in FFHQ-256, LSUN Church-256, and LSUN Horse-256 datasets.
         Each column corresponds to samples generated from the same noise vector $z$.
         Averaged L1 distances between 10K samples from teacher and student are reported below each method.
         The lowest distortion of the proposed method validates that ours has enhanced capability to preserve the diversity in the image space.
        % cagan*, stylekd*, ours L1 기반 차이 10000장 비교
        % FFHQ-256 0.2346, 0.2396, 0.1972
        % CHURCH-256 0.2591, 0.2919, 0.2528
        % HORSE-256 0.2813, 0.2953, 0.2694
        % MEAN 0.2583 0.2756 0.2398
     }
     \label{qualitative}
    \vspace{-0.5cm}
\end{figure*}

\begin{table}
\small
  \caption{
      \textbf{Quantitative results on StyleGAN3.}
      Comparison with recent baselines~\cite{liu2021content, xu2022mind} in FFHQ dataset.
      }
      \vspace{-0.2cm}
    \centering
    \begin{tabular}{c|ccc} \toprule 
    Metric &FID~$\downarrow$ &P~$\uparrow$ &R~$\uparrow$ \\ \midrule \midrule
    Teacher &4.76 &0.647 &0.484 \\ \midrule
    CAGAN~\cite{liu2021content} &14.25 &0.636 &0.272 \\
    StyleKD~\cite{xu2022mind} &9.57 &0.647 &0.376 \\
    Ours &\textbf{8.40} &\textbf{0.653} &\textbf{0.397} \\
    \bottomrule
    \end{tabular}
    \vspace{-0.4cm}
\label{stylegan3}
\end{table}

\subsection{Qualitative Results}
\noindent\textbf{Generated samples from same noise input.}
For qualitative comparison, we show the samples generated by different compressed models from the same noise vector $z$ in Fig.~\ref{qualitative}. 
As noted in StyleKD~\cite{xu2022mind}, it would be desirable for a compressed network to reconstruct the sample generated by the teacher network.
However, we observed that the baseline methods struggle to produce a sample inheriting the characteristics of images from the teacher network. 
For instance, CAGAN$^{*}$ and StyleKD$^*$ generate face images with different shapes of eyes in the third and fourth column of Fig.~\ref{qualitative}.
Similarly, in the first column of LSUN Horse, horses generated by baselines hardly maintain the brown and white color patterns available in the teacher's image, while the proposed method successfully synthesizes those patterns.
These results show that the channel pruning with the consideration of the latent space and its effect on images leads to better diversity preservation of generated images.

\subsection{Ablation Studies and Analysis}
We further discuss the influence of the proposed components and other hyperparameters by comprehensive ablation studies. For this purpose, we train the ablated models with different configurations in FFHQ-256.

\vspace{0.2cm}
\noindent\textbf{Number of directions and strength for perturbations.}
We investigate the number of directions ($N$) and the strength of perturbation ($\alpha$). Due to the computational costs, we report the mean and standard deviations of the best FID scores up to 50K iterations over 3 times trials. As reported in Tab.~\ref{ablation_param}-(a), our method achieves the lowest FID score when $N=10$. Additionally, we explore the impact of the strength of perturbation ($\alpha$). While $\alpha=5$ yields the best FID in this scenario, we consider both $\alpha=5,10$ as viable options.

\begin{table}[t]
\centering
\small
\vspace{-0.2cm}
\caption{
  \textbf{Ablation on the perturbation parameters.}
  In this table, we report mean and standard deviation of FID$_{\text{early}}$ measured from the models training until 50K iterations 3 times.  We denote $N$ and $\alpha$ as the number of perturbations for each original latent vector and the strength parameter of perturbations, respectively.
  \textbf{Ablation on the direction sampling.}
    For comparing semantic and random directional vectors, we utilize the GANSpace~(PCA) method to obtain the semantic directional vector. The FID score with semantic directional vectors is slightly better than that of the randomly sampled vectors.
}
    \begin{subtable}[b]{\columnwidth}
    \centering
        \begin{tabular}{c|c|c|c}\toprule
        $N$ &FID$_{\text{early}}$~$\downarrow$ &$\alpha$ &FID$_{\text{early}}$~$\downarrow$ \\ \midrule \midrule
        $N=5$ &12.46 ± 0.2 &$\alpha=1$ &13.50 ± 0.3   \\
        $N=10$ &\textbf{12.08 ± 0.3} &$\alpha=5$ &\textbf{12.08 ± 0.3} \\
        $N=20$ &12.47 ± 0.2 &$\alpha=10$ &12.09 ± 0.2 \\
        \bottomrule
        \end{tabular}
        \centering
        \caption{
        Ablation on the perturbation parameters
        }
    \end{subtable}
    \begin{subtable}[b]{0.8\columnwidth}
    \centering
    \begin{tabular}{c|ccc}\toprule
    Direction &FID~$\downarrow$ &P~$\uparrow$ &R~$\uparrow$ \\ \midrule \midrule
    Random &6.50 &0.692 &\textbf{0.342} \\
    PCA &\textbf{6.35} &\textbf{0.706} &0.339 \\
    \bottomrule
    \end{tabular}
    \centering
    \caption{Ablation on the perturbation directions}
    \end{subtable}
    \label{ablation_param}
    \vspace{-0.4cm}
\end{table}

\begin{table}[t]
\centering
\small
\caption{\textbf{Ablation on the type of the importance scores.}
  Comparison of FID score between the average-based importance score~($S^{\mu}$) and the proposed diversity-sensitive importance score~($S^{\sigma}$) in FFHQ-256 and LSUN Horse datasets.}
  \vspace{-0.2cm}
    \setlength\tabcolsep{4.0pt}
    \begin{tabular}{c|ccc} \toprule 
    Dataset &StyleKD* &Ours~(w/~$S^{\mu}$) &Ours~(w/~$S^{\sigma}$) \\ \midrule \midrule
    FFHQ-256 &7.47 &6.71 &\textbf{6.35} \\
    LSUN Horse &5.69 &5.36 &\textbf{5.11} \\
    \bottomrule
    \end{tabular}
    \centering
    \label{ablation_score}
    \vspace{-0.4cm}
\end{table}

\vspace{0.2cm}
\noindent\textbf{Latent perturbations by random vs. PCA-based directions.}
We conduct the ablation study on directional vectors for latent perturbations.
Specifically, we compare the performance of directional vectors derived from GANSpace~\cite{harkonen2020ganspace} and vectors sampled from a random normal distribution~($d_{\text{rand}}\sim\mathcal{N}(0,1)$).
Tab.~\ref{ablation_param}-(b) shows a slight degradation in performance when using random directional vectors for the perturbation. 
However, this degradation is negligible, and our pruned model still outperforms compared to baselines~\cite{xu2022mind, liu2021content} in terms of FID metric.

Thus, this finding highlights the significance of perturbation itself regardless of the type (either random or PCA) of perturbation in capturing sample diversity.
Nevertheless, semantic directional vectors from GANSpace yield a lower FID compared to randomly sampled vectors. This result motivates us to select GANSpace-based directions as the default configuration for the latent vector perturbations.

\begin{figure}[t!]
    \centering
    \vspace{-0.25cm}
     \begin{subfigure}[b]{0.48\textwidth}
         \centering
         \includegraphics[width=\textwidth]{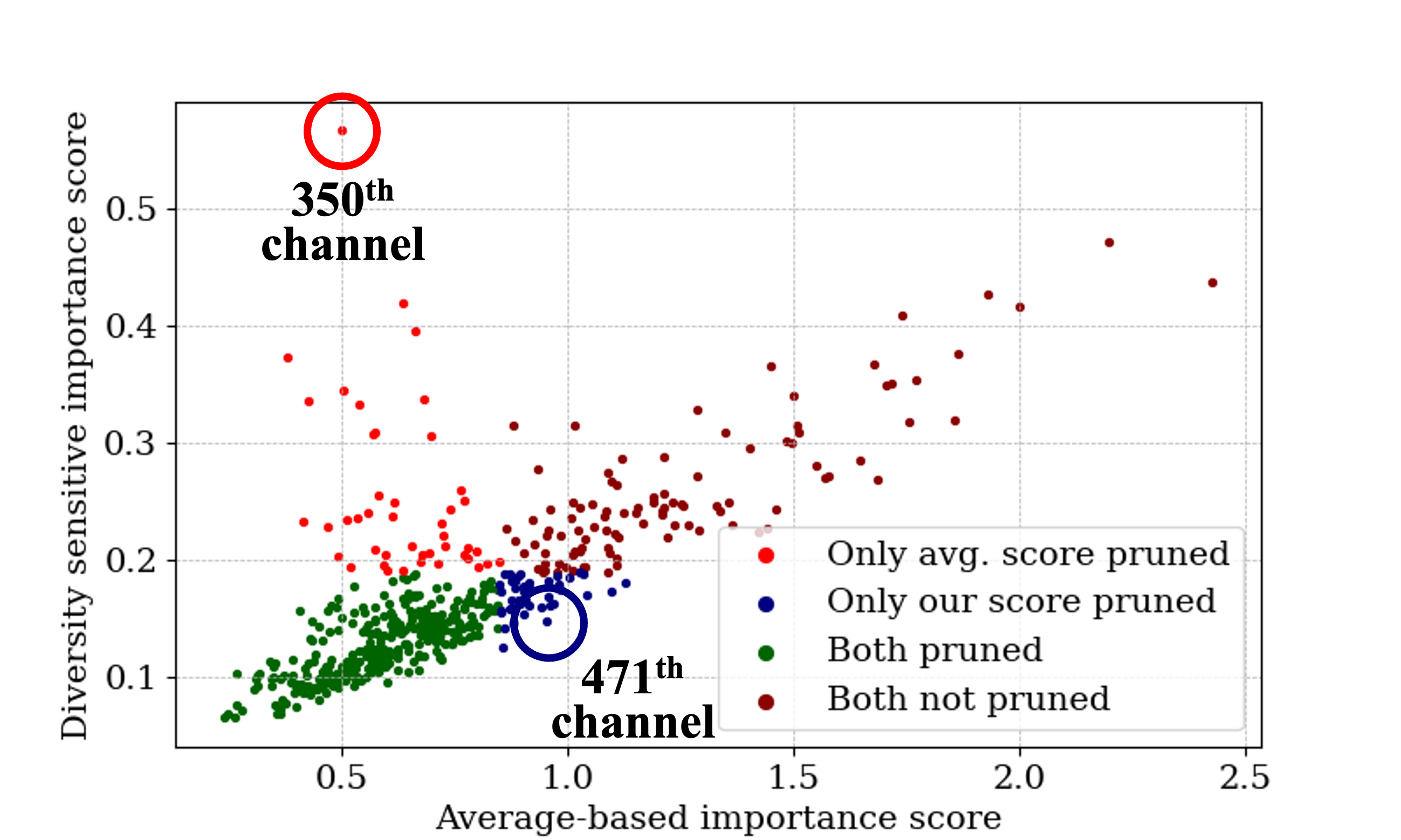}
         \vspace{-0.4cm}
         \caption{Channel scatter plot for avg. and our diversity-sensitive scores~($S^\mu,S^\sigma$)}
         \label{fig_scatter}
     \end{subfigure}
     \begin{subfigure}[b]{0.21\textwidth}
         \centering
         \includegraphics[width=\textwidth]{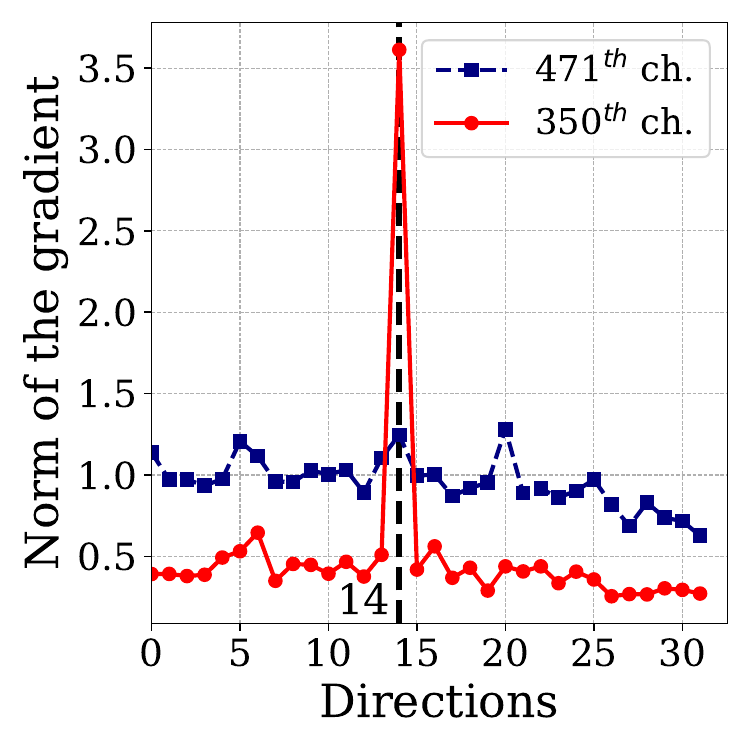}
         \vspace{-0.6cm}
         \caption{Norm of the grad. of each dir.}
         \label{fig_gradnorm}
     \end{subfigure}
    \hspace*{-0.2cm}
     \begin{subfigure}[b]{0.265\textwidth}
         \centering
         \includegraphics[width=\textwidth]{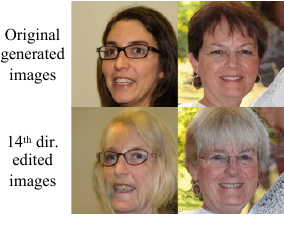}
         \vspace{-0.6cm}
         \caption{Edited samples for $14^\text{th}$ direction}
         \label{fig_age}
     \end{subfigure}
     \vspace{-0.5cm}
     \caption{
     \textbf{Specific examples for pruned channels}
         (a)~We provide a scatter plot of $S^{\mu}$ and $S^{\sigma}$ scores of all channels in 6${^\text{th}}$ layer of teacher generator trained on FFHQ-256.
        \textcolor{brown}{Brown} and \textcolor{green}{green} dots represent channels that are always pruned and not pruned, respectively. \textcolor{blue}{Blue} and \textcolor{red}{Red} dots indicate channels that only survive in $S^{\mu}$ and $S^{\sigma}$, respectively.
        (b)~The 350${^\text{th}}$ channel exhibits high sensitivity to the 14${^\text{th}}$ direction from PCA.
        (c)~The 14${^\text{th}}$ direction corresponds to an age-related perturbation. The $S^{\mu}$ score prunes the 350${^\text{th}}$ channel, while the $S^{\sigma}$ score preserves this channel, which demonstrates high sensitivity to age variation.
        This result shows that $S^{\sigma}$ aims to retain semantic image changes compared to $S^{\mu}$.}
    \label{fig_channel_score}
    \vspace{-0.4cm}
\end{figure}

\vspace{0.2cm}
\noindent\textbf{Type of the importance score measurement.}
We conduct the ablation study to validate the effectiveness of our diversity-sensitive importance score. As shown in Tab.~\ref{ablation_score}, the proposed diversity-sensitive importance score~($S^{\sigma}$) outperforms the average-based importance score~($S^{\mu}$) in terms of FID. Specifically, the proposed diversity-sensitive method provides 6.35 FID, improving 0.36 points over the average-based score.
Similar trends can be found in LSUN Horse dataset. These results confirm that suppressing the importance score derived from latent code itself is crucial.

% \paragraph{Specific examples for pruned channels}
\vspace{0.2cm}
\noindent\textbf{Specific examples for pruned channels}
In Fig~\ref{fig_channel_score}, we examine the score of each channel in the teacher generator using a scatter plot, with $S^\mu$ on the x-axis and $S^\sigma$ on the y-axis. As visualized in Fig.~\ref{fig_channel_score}~(a), $S^\mu$ and $S^\sigma$ of most channels are linearly correlated. However, upon closer examination of outliers, a significant portion of these outliers possess low $S^\mu$ values but high $S^\sigma$ values.
These outlier channels generally contribute to a smaller number of directions but are highly activated for particular directions.
To illustrate this, we visualize the gradient norms for 350$^\text{th}$ and 471$^\text{th}$ channels in Fig.~\ref{fig_channel_score}~(b). The 350$^\text{th}$ channel exerts a significant influence on the 14$^\text{th}$ direction. This particular direction corresponds to notable image differences, such as age-related perturbations, as shown in Fig.~\ref{fig_channel_score}~(c). Thus, our $S^\sigma$ can assign higher importance scores for these outliers.

\vspace{-0.1cm}
\section{Conclusion}
In this work, we address the issue of a compressed StyleGAN generator struggling to preserve the sample diversity of teacher, observed in previous GAN compression methods. To alleviate this problem, we propose a simple yet effective channel pruning method that calculates channel importance scores based on their sensitivity in the image space according to perturbations in latent space. 
Extensive experiments demonstrate that our pruning method significantly enhances sample diversity, validating its superior performance in terms of FID.

\vspace{-0.1cm}
\section*{Acknowledgments}
\vspace{-0.2cm}
This work was supported in part by MSIT/IITP (No. 2022-0-00680, 2019-0-00421, 2020-0-01821, 2021-0-02068), and MSIT\&KNPA/KIPoT (Police Lab 2.0, No. 210121M06).

{
    \small
    \bibliographystyle{ieeenat_fullname}
    \bibliography{main}
}

\clearpage
\setcounter{page}{1}
\maketitlesupplementary

\section{Additional Results}

\vspace{0.2cm}
\noindent\textbf{Details about the datasets.}
We utilize three popular datasets in the research field of GAN: FFHQ, LSUN Church, LSUN Horse. FFHQ~\cite{karras2019style} contains $70,000$ human facial images with a resolution of $1024\times 1024$, and we also test on its resized version with a resolution of $256\times 256$. LSUN Church~\cite{yu2015lsun} comprises $126,227$ outdoor church images with a resolution of $256\times 256$, and LSUN Horse~\cite{yu2015lsun} includes $2$ million horse images with a resolution of $256\times 256$. However, in alignment with the experimental setting of StyleGAN2, we only utilize a subset of 1 million images from the LSUN Horse dataset.

\begin{figure*}[t]
    \centering
     \includegraphics[width=0.9\textwidth]{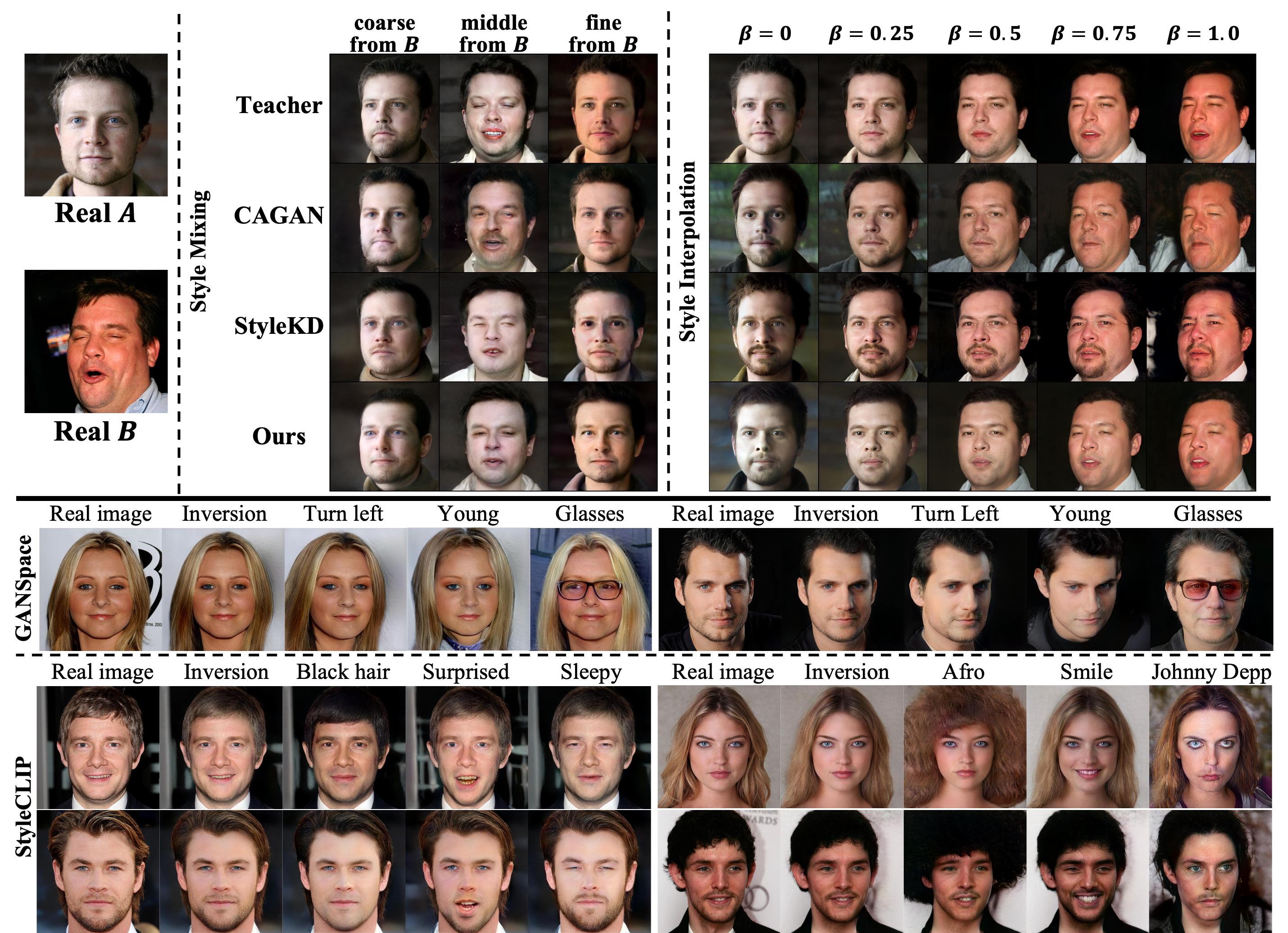}         
     \caption{
         \textbf{Real image editing results by our model}.
         (Upper part) middle: style mixing, right: interpolation, and (Lower part) GANSpace and StyleCLIP examples.
         These results validate that our compressed GAN can be applied to real-world tasks.
     }
     \label{editing}
\end{figure*}

\vspace{0.2cm}
\noindent\textbf{Image editing results of our compressed generator.}
We perform various real-world tasks, including style mixing, style interpolation, and latent editing using GANSpace~\cite{harkonen2020ganspace} and StyleCLIP~\cite{patashnik2021styleclip}, as shown in Fig.~\ref{editing}.

For style mixing, we define coarse, middle, and fine layers as [0:2]$^{\text{th}}$, [4:7]$^{\text{th}}$, and [9:12]$^{\text{th}}$ layers, respectively, following baseline~\cite{xu2022mind}. In this process, we inject each level of latent code from image B into image A. As in fine layer injection, our model successfully transplants the tone from image B while preserving the identity of image A.

For style interpolation, we perform linear interpolation between the inverted latent codes~($w_A, w_B$) to generate style-interpolated images; $w_\text{interp}=w_A\times(1-\beta)+w_B\times\beta, \beta\in[0,1]$.
StyleKD shows a glasses artifact in the interpolated image despite neither image A nor B wearing the glasses. Another baseline, CAGAN shows low-quality inversion and interpolated images. In contrast, our model shows high-quality editing results that are similar to the teacher model.

For editing via GANSpace, we identify the important latent directions using PCA on 50,000 randomly sampled $w$. Then, we edit the inverted real-world images using the computed latent directions. As a result, we found 2$^{\text{th}}$, 4$^{\text{th}}$, and 9$^{\text{th}}$ latent direction captures the attributes ``Turn left", ``Young", and ``Glasses", and they are successfully applicable for editing in the compressed generator. Furthermore, we also confirmed that these directions are shared their semantic changes regardless of latent code $w$.

Furthermore, we validate the suitability of the proposed method in real-world applications, text-driven image editing. For experiments, we adopt the StyleCLIP~\cite{patashnik2021styleclip} as editing method.
As shown in Fig.~\ref{editing}, we observe that the compressed generator successfully works for both tasks, inverting and editing on the given real-world images and various input text prompts.
These experimental results confirm the generative capability of our compression technique and provide strong evidence for the practical applicability of our compressed generator.

\vspace{0.2cm}
\noindent\textbf{Actual speed gains.}
We measure the inference time per image for synthesizing 1,000 images (with a batch size of 4) on a single RTX 3070 GPU. The results are presented in the Tab.~\ref{actual_speed}.
In real-world scenarios, our pruning model~($p_r=0.7$) achieves a substantial speedup of 2.35x faster in 256 resolution and an impressive 3.76x faster in 1024 resolution compared to the teacher model. We did not include the inference speeds of other baselines since the number of parameters and FLOPs of our model is identical to theirs.
These results highlight the significant computational efficiency gained through our pruning approach, showing its practical properties in real-world applications.

\begin{table}
\centering
\caption{\textbf{Actual speed comparison with teacher model.}
Our compressed model significantly accelerates inference speed compared to the teacher model.}
\begin{tabular}{|c|c|c|}
\hline
Inference Time~(ms) & Teacher & Ours  \\ \hline
FFHQ-256            & 12.90   & 5.48~\textcolor{red}{(2.35x)}  \\ \hline
FFHQ-1024           & 45.23   & 12.03~\textcolor{red}{(3.76x)}  \\ \hline
\end{tabular}
\label{actual_speed}
\vspace{-0.4cm}
\end{table}

\begin{figure*}[t]
    \centering
     \includegraphics[width=0.9\textwidth]{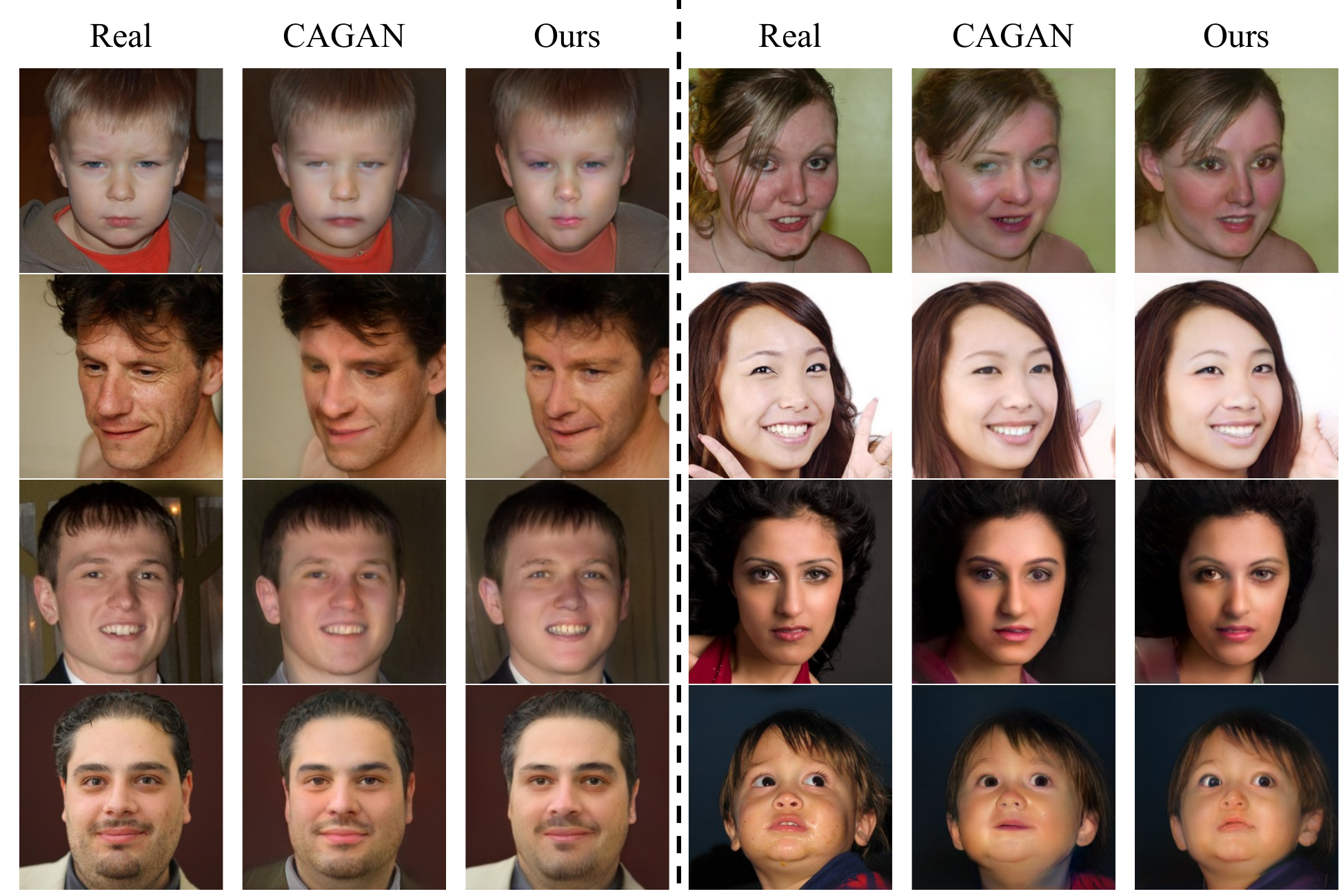}         
     \caption{
         \textbf{Projection examples of the real-world dataset.}
         We visualize the real-world images from Helen-Set55~\cite{liu2021content} and its projected results.
         We use the generators trained on FFHQ-1024 datasets.
         The visual similarity between the real image and the projected one shows that our compressed models have sufficient capability to express real samples.
         Note that, these real images are not included in our training dataset.
     }
     \label{projection}
     \vspace{-0.4cm}
\end{figure*}

% \paragraph{Projection examples of the real-world dataset.}
\vspace{0.2cm}
\noindent\textbf{Projection examples of the real-world dataset.}
We provide the projection results for real samples from Helen-Set55~\cite{liu2021content}.
Note that, these real-world images are not included in our training dataset.
Models trained on FFHQ-1024 datasets are used for this experiment.
As shown in Fig.~\ref{projection}, we verify that our compressed models are able to capture sufficient information about the real samples to reconstruct them.

% \paragraph{Additional comparison for sample diversity.}
\vspace{0.2cm}
\noindent\textbf{Additional comparison for sample diversity.}
Diversity refers to the generator's capacity to produce varied images. To assess this, we begin by sampling 5,000 images from identical random latent vectors $z$ for each model trained by FFHQ-256 dataset. Next, we select a reference sample and identify its nearest neighbors among the other sampled images, by measuring the L2 distance between these images. A smaller L2 distance implies the generator is producing similar images, whereas a larger value signifies diverse image generation. This process is repeated for all 5,000 samples, enabling us to provide not only the minimum distance but also the average and maximum distances for a comprehensive assessment of diversity.

Our model shows the largest distance between samples among compressed GANs, as shown in Tab.~\ref{diversity}.
It implies that our model generates more diverse samples from distinct latent vectors.

\begin{table}
\centering
\caption{
\textbf{Minimum and average L2 average distance of each generated image between the other generated images.}
Please refer to the detailed calculations within the paragraph.}
\begin{tabular}{|c|c|c|}
\hline
        & Minimum    & Average    \\ \hline
Teacher & 0.0449 & 0.1321 \\ \hline
Ours    & \textbf{0.0437} & \textbf{0.1313} \\
StyleKD & 0.0427 & 0.1303 \\ 
CAGAN   & 0.0424 & 0.1304 \\ \hline
\end{tabular}
\label{diversity}
\vspace{-0.4cm}
\end{table}

\begin{figure*}
    \centering
     \includegraphics[width=0.95\textwidth]{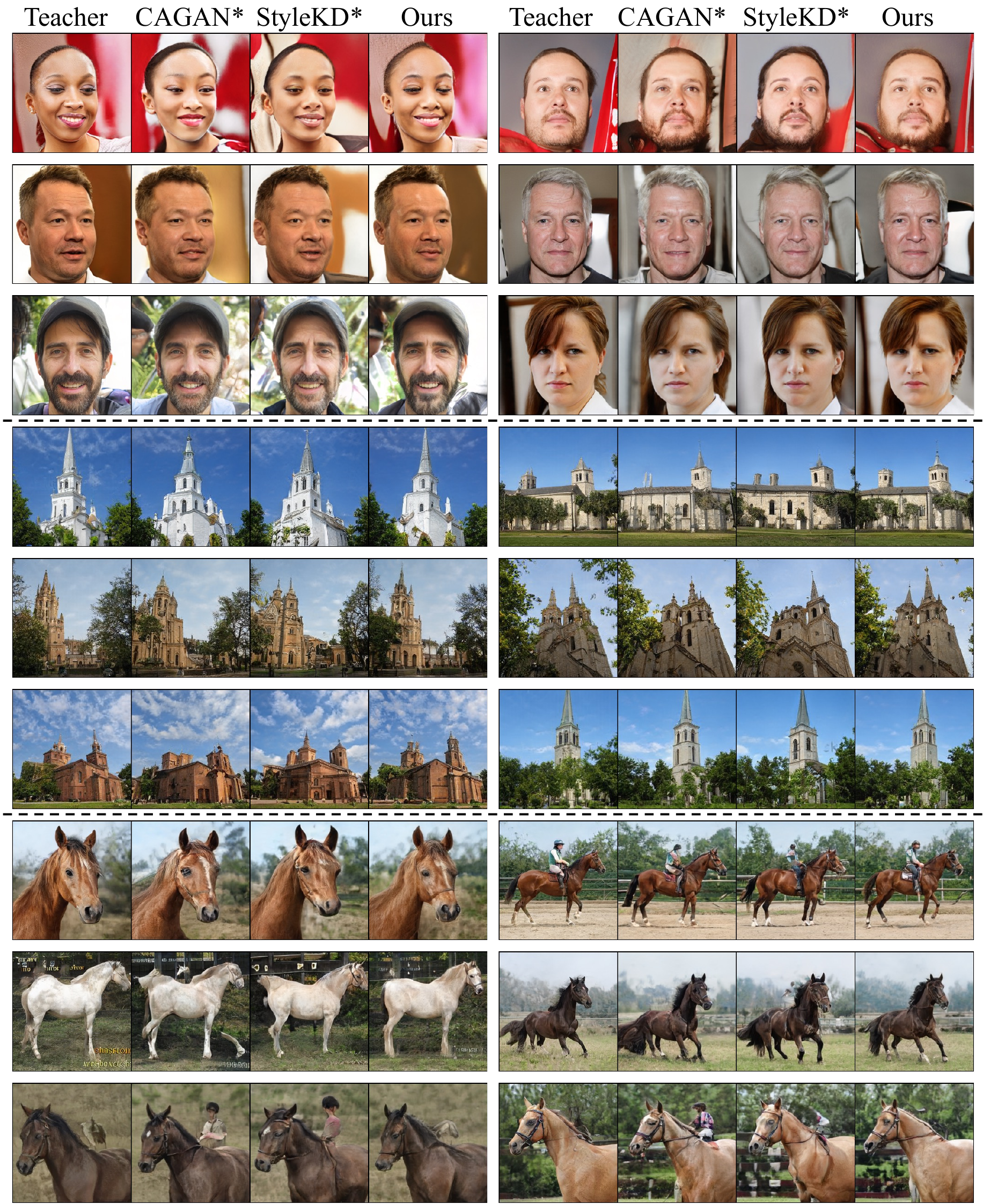}         
     \caption{
         \textbf{Qualitative comparison with baselines on various datasets.}
         For comparison, we visualize the generated samples from ours and baselines~\cite{liu2021content,xu2022mind} in FFHQ-256, LSUN Church, and LSUN Horse datasets.
         Each half of the row corresponds to samples generated from the same noise vector $z$.
     }
     \label{256_qual}
\end{figure*}

\begin{figure*}
    \centering
     \includegraphics[width=0.85\textwidth]{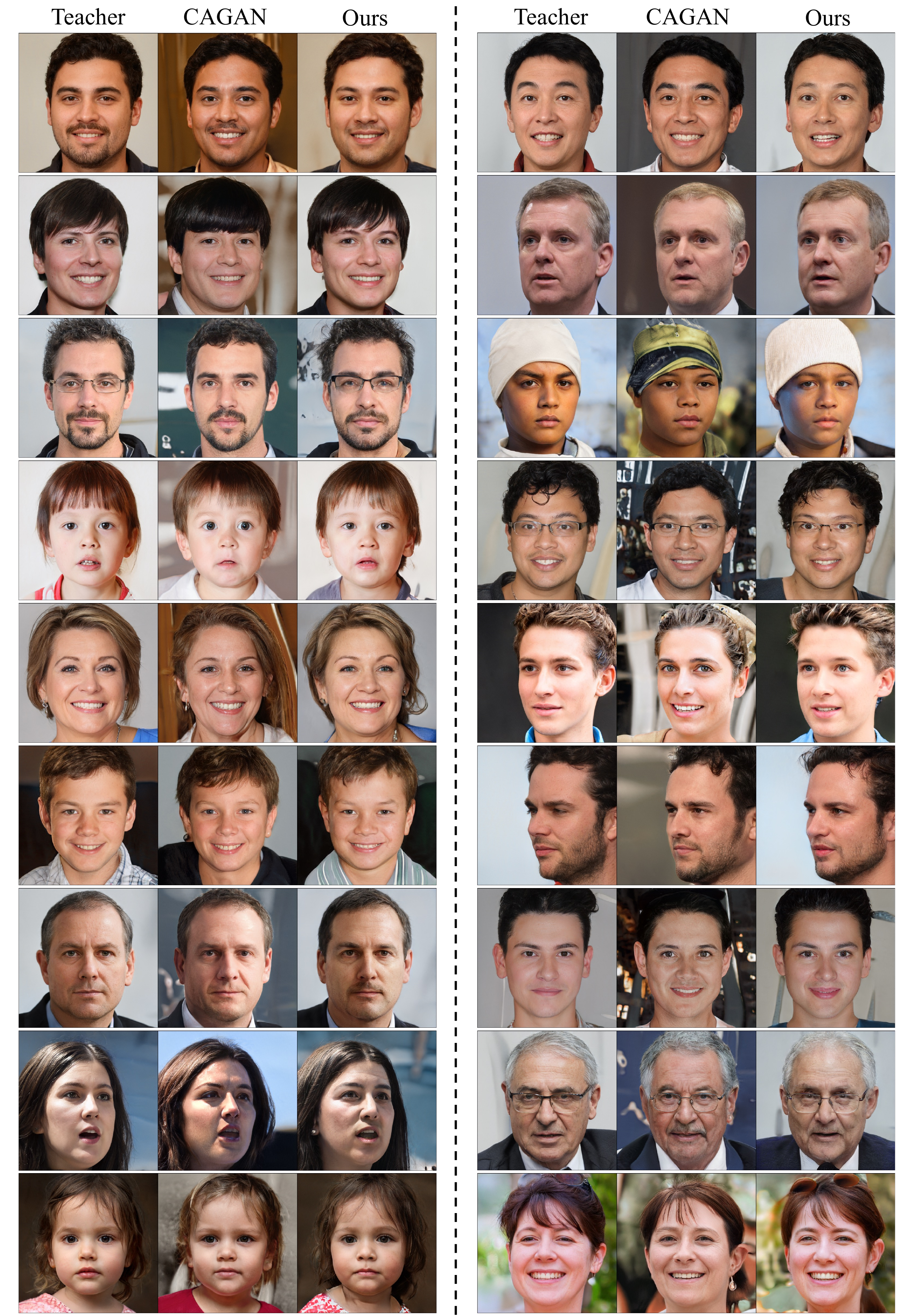}         
     \caption{
         \textbf{Qualitative comparison with baselines on the high-resolution dataset.}
         For comparison, we visualize generated samples from ours and baselines~\cite{liu2021content,xu2022mind} in FFHQ-1024 dataset.
        Each half of the row corresponds to samples generated from the same noise vector $z$.
     }
     \label{1024_qual}
\end{figure*}

% \paragraph{Additional generated samples from same noise input.}
\vspace{0.2cm}
\noindent\textbf{Additional generated samples from same noise input.}
We provide an additional visual comparison between the proposed method and baselines~\cite{liu2021content,xu2022mind}.
Specifically, we synthesize the samples from the identical noise input for every methods on four datasets; FFHQ-256, FFHQ-1024, LSUN Church, and LSUN Horse.
As a result, the generated samples from ours are more visually similar to samples from the teacher model compared to baselines, and this result is achieved consistently regardless of the type of dataset and its resolution, as shown in Fig.~\ref{256_qual} and Fig.~\ref{1024_qual}.
Our method exhibits a high similarity to the teacher-generated images compared to the baselines. As a result, our model demonstrates an enhanced capability to preserve sample diversity of the teacher model.

\begin{figure*}
    \centering
     \includegraphics[width=0.85\textwidth]{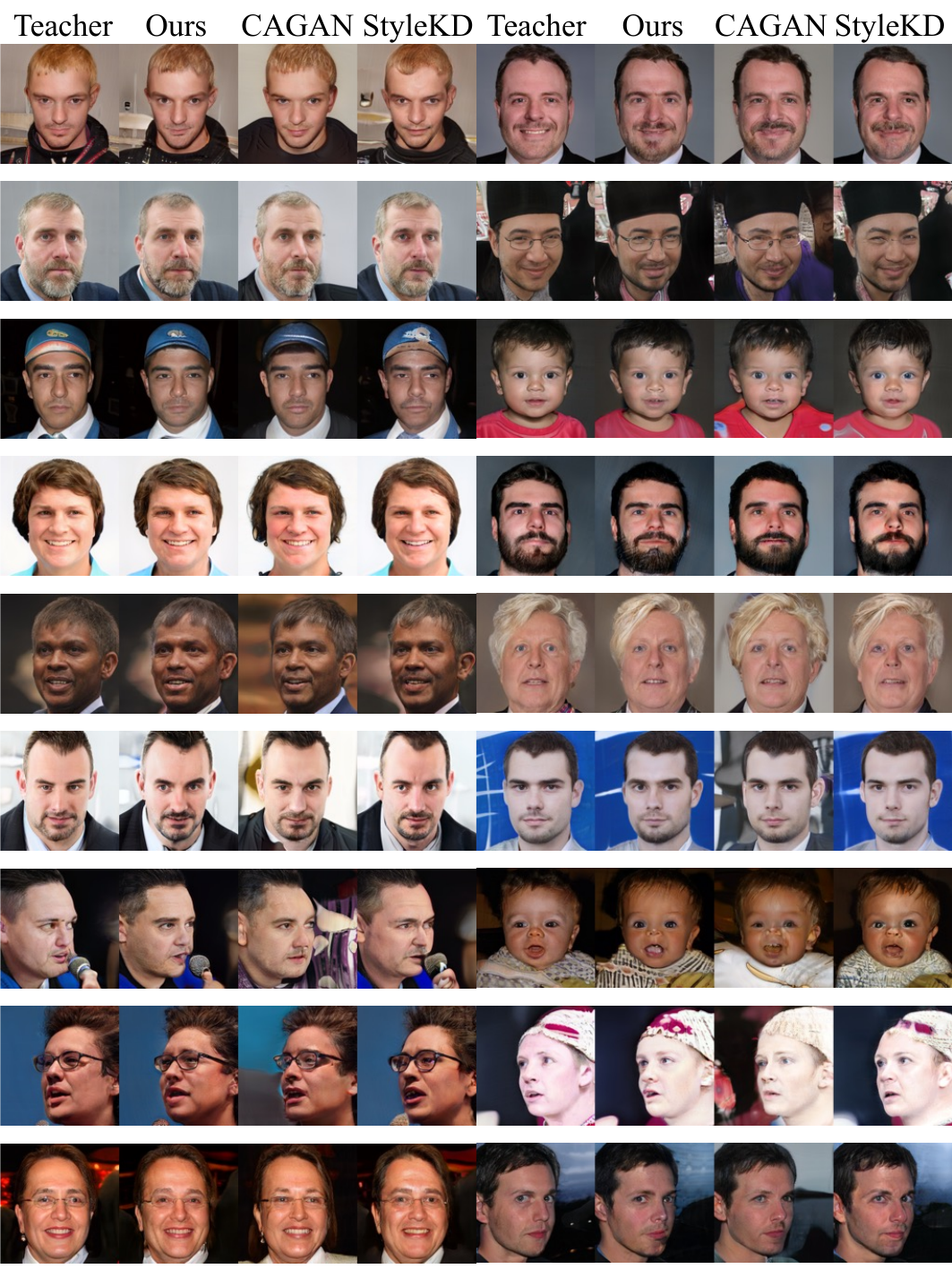}         
     \caption{
         \textbf{Qualitative comparison with baselines on StyleGAN3.}
         For comparison, we visualize generated samples from ours and baselines~\cite{liu2021content,xu2022mind} on StyleGAN3-T~\cite{karras2021alias} in the FFHQ-256 dataset.
        Each half of the row corresponds to samples generated from the same noise vector $z$.
     }
     % ours 0.1582, stylekd : 0.1658, cagan : 0.2664
     \label{sg3_qual}
\end{figure*}

\section{Further Analysis}

\begin{figure*}
    \centering
     \includegraphics[width=0.85\textwidth]{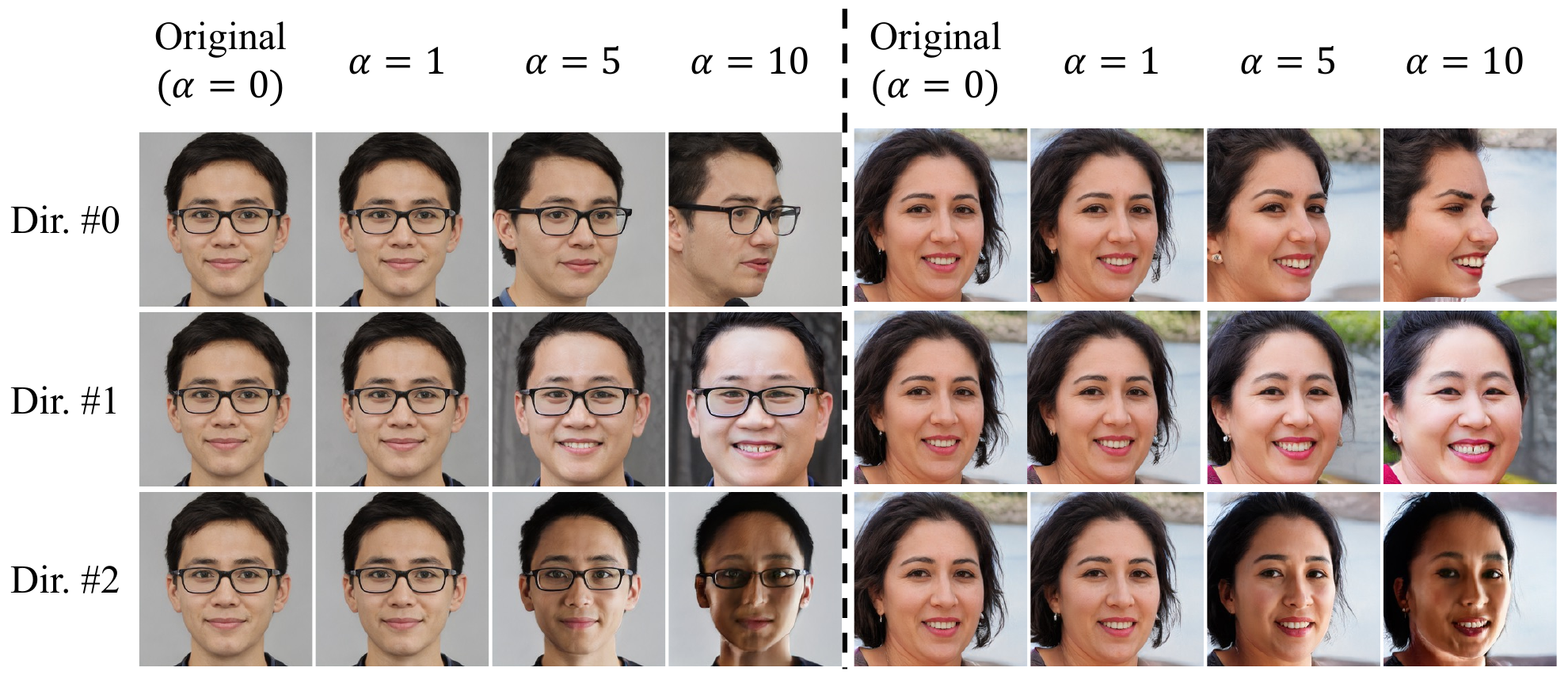}         
     \caption{
         \textbf{Visual transition as strength of perturbation changes. }
         We first obtain three directional vectors~(Dir.) using GANSpace~\cite{harkonen2020ganspace}.
         Next, we generate the samples and their perturbed counterparts with the different strength parameters for the perturbations ($\alpha$).
         When $\alpha=1$, perturbed samples only show minor changes to identify the image difference that latent changes lead to.
         Conversely, when $\alpha=5,10$, the perturbed samples show sufficient pixel-level differences to detect the effect of latent variations.
         Similarly, our ablation study~(Sec.~4.4 in main manuscript) validates that $\alpha=5,10$ are the proper strength of perturbations by achieving lower FID$_{\text{early}}$ metric.
         Therefore, when selecting the strength of perturbation changes, it is crucial to ensure sufficient pixel-level differences in the perturbed samples to effectively capture the effects of latent variations.
     }
     \label{alpha}
\end{figure*}

\noindent\textbf{Visual transition as strength of perturbation changes.}
We visually analyze the generated samples along with their perturbed counterparts generated from various strength parameters of the perturbations ($\alpha$).
As shown in Fig.~\ref{alpha}, selecting $\alpha=5,10$ yields the perturbed samples with a sufficient magnitude of pixel-level changes in images.
This observation aligns with our ablation study~(``Ablation'' section in main manuscript), which demonstrates that $\alpha=5,10$ is an adequate strength for perturbations as supported by FID$_{\text{early}}$ metric.
Specifically, our ablation study denotes that the FID$_{\text{early}}$ metric values for $\alpha=5,10$ (12.08 and 12.09, respectively) are lower than $\alpha=1$ (13.50).
Therefore, when selecting the strength of perturbation changes, it is crucial to ensure that there are significant visual transitions in the perturbed samples to effectively capture the effects of latent perturbations.

\begin{table}
\footnotesize
\centering
\caption{Comparison on FastGAN~($p_r=0.5$, 60K iters, 3 times)}
\vspace{-0.2cm}
\begin{tabular}{l|c|ccc}
\toprule
FID~$\downarrow$ & Teacher & Scratch & StyleKD & Ours \\ 
\midrule
Dog~(256) & 52.05 ± 0.2 & 56.34 ± 0.3 & 55.10 ± 0.7 & \textbf{53.17 ± 0.3} \\
\bottomrule
\end{tabular}
\vspace{-0.3cm}
\label{tab_reb_fast}
\end{table}

% \paragraph{More experiments in other network structure~(FastGAN).}
\vspace{0.2cm}
\noindent\textbf{More experiments in other network structure~(FastGAN).}
We conduct the other GAN architecture, FastGAN~\cite{liu2020towards}, distinct from the structure of StyleGAN.
As shown in Tab.~\ref{tab_reb_fast}, proposed method shows superior performance even in a different network structure, validating the generalizability of the proposed method.

\begin{table}
\centering
\footnotesize
\caption{Ablation study of the image difference metric~(220K iters)}
\vspace{-0.2cm}
\begin{tabular}{l|cc|l|cc}
\toprule
Dataset & L1 & LPIPS & Dataset & L1 & LPIPS \\ 
\midrule
FFHQ-256 & \textbf{7.05} & 7.32 & Horse-256 & 6.49 & \textbf{6.18} \\
\bottomrule
\end{tabular}
\vspace{-0.3cm}
\label{tab_reb_lpips}
\end{table}

\vspace{0.2cm}
\noindent\textbf{Utilization of LPIPS for capturing image difference.}
To investigate the effects of distance measure, we additionally conduct experiment that prunes channels with LPIPS as image difference. 
As reported in Tab.~\ref{tab_reb_lpips}, we observe that two distance measures~(L1, LPIPS) perform similarly.
We hypothesize that semantic perturbation we used~(i.e. PCA directions) already encourages the model to be aware of semantics, although guidance consists of pixel-level L1 distance.

\begin{figure*}[t!]
    \vspace{-0.2cm}
\includegraphics[width=0.80\textwidth]{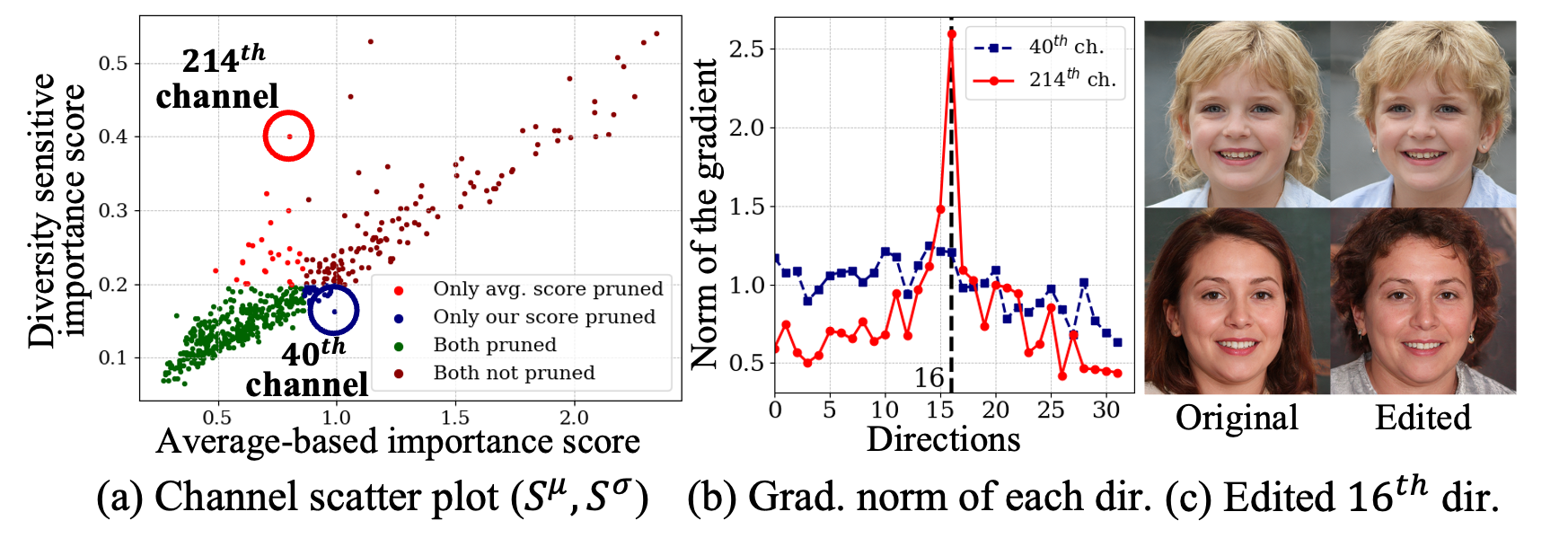}
    \centering
    \vspace{-0.3cm}
    \caption{
     \textbf{Additional examples for pruned channels}
        (a)~We provide additional scatter plot as same as Fig.~5 in main manuscript.
        (b)~The 214${^\text{th}}$ channel exhibits high sensitivity to the 16${^\text{th}}$ direction from PCA.
        (c)~The 16${^\text{th}}$ direction corresponds to an hair length related perturbation. The $S^{\mu}$ score prunes the 214${^\text{th}}$ channel, while the $S^{\sigma}$ score preserves this channel, which demonstrates high sensitivity to the hair length variations.}
    \vspace{-0.3cm}
    \label{fig_reb_scatter}
\end{figure*}

\vspace{0.2cm}
\noindent\textbf{More examples for pruned and not-pruned channels between the $S^\mu$ and $S^\sigma$ scores.}
In Fig.~\ref{fig_reb_scatter}, we further visualize channels in 5$^\text{th}$ layer following the experimental settings of Fig.~5 in main manuscript.
The 214$^\text{th}$ channel exhibits low $S^\mu$ and high $S^\sigma$ values with strong activation for 16th direction, associating with hair length.

\begin{table}
\centering
\small
\caption{Quantitative comparisons with baselines~($p_r=0.5$, 100K iters, FFHQ-256 dataset)}
\vspace{-0.2cm}
\begin{tabular}{l|ccc}
\toprule
 & Ours & StyleKD & CAGAN \\ 
\midrule
FID~$\downarrow$ & \textbf{6.78} & 8.79 & 11.68 \\
\bottomrule
\end{tabular}

\vspace{-0.3cm}
\label{tab_reb_msg}
\end{table}

% \paragraph{Comparison with a different pruning ratio}
\vspace{0.2cm}
\noindent\textbf{Comparison with a different pruning ratio.}
We train ours and baseline methods up to 100K iters with $p_r=0.5$. The proposed method demonstrates superior performance compared to the baselines, as shown in Tab.~\ref{tab_reb_msg}.

\vspace{0.2cm}
\noindent\textbf{An overview of our method's implementations.}
Our pruning implementation follows the stages outlined below:

1. Prepare the teacher model ($f,g$), along with the perturbation vector $d$.

2. Sample a latent vector $w$ and its perturbed counterpart $(w+\alpha d)$.

3. Generate two images $g(w)$ and $g(w+\alpha d)$, and perform backpropagation from the loss $\mathcal{L}_{\text{diff}}=\vert g(w)-g(w+\alpha d) \vert$.

4. Accumulate the gradients $\textbf{G}_{\text{perturb}}$.

5. For 1,000 iterations, repeat steps 2 to 4.

6. Calculate the diversity-sensitive importance score $S^{\sigma}$ and prune channels based on this score.

\vspace{0.2cm}
\noindent\textbf{Broader Impacts.}
In today's social media landscape, the generation of fake images of celebrities or sports stars using generative models is a major concern. Our proposed compression method not only address the computational challenges but also brings attention to the potential misuse of such techniques.
To mitigate the negative impact of fake images, detection models~\cite{heo2023deepfake,9412711} offers a solution to minimize the harm caused by these fake images. It is crucial to also consider the ethical implications of such technologies and promote responsible use to prevent malicious exploitation.

\vspace{0.2cm}
\noindent\textbf{Limitations.}
The proposed pruning method aims to preserve the sample diversity of teacher network as much as possible. Hence, the samples that can potentially disturb the effective transfer of knowledge of teacher~(e.g. samples with degenerated quality) also can be preserved in the student network. This may hinder the further improvement in the generation performance of student network.

\end{document}